\documentclass{article}
\usepackage{spconf,amsmath,graphicx,hyperref}
\usepackage{multirow}
\usepackage{multicol}
\usepackage{booktabs}
\usepackage{makecell}
\usepackage{CJKutf8}
\usepackage{tablefootnote}
\usepackage{subfigure}
\usepackage{graphicx}
\usepackage{booktabs}
\usepackage{enumitem}
\title{LESS: Large Language Model Enhanced Semi-Supervised Learning for Speech Foundational Models Using in-the-wild Data}
\name{Wen Ding, Fan Qian}
\address{NVIDIA Corporation\\
\url{{wend, fqian}@nvidia.com}}
\begin{document}
% \ninept
%
\maketitle
\begin{abstract}
Although state-of-the-art Speech Foundational Models can produce high-quality text pseudo-labels, applying Semi-Supervised Learning (SSL) for in-the-wild real-world data remains challenging due to its richer and more complex acoustics compared to curated datasets. To address the challenges, we introduce LESS (Large Language Model Enhanced Semi-supervised Learning), a versatile framework that uses Large Language Models (LLMs) to correct pseudo-labels generated on in-the-wild data. In the LESS framework, pseudo-labeled text from Automatic Speech Recognition (ASR) or Automatic Speech Translation (AST) of the unsupervised data is refined by an LLM, and further improved by a data filtering strategy. Across Mandarin ASR and Spanish-to-English AST evaluations, LESS delivers consistent gains, with an absolute Word Error Rate reduction of 3.8\% on WenetSpeech, and BLEU score increase of 0.8 and 0.7, achieving 34.0 on Callhome and 64.7 on Fisher testsets respectively. These results highlight LESS’s effectiveness across diverse languages, tasks, and domains. We have released the recipe as open source to facilitate further research in this area. 
% Ablation studies conducted with various LLMs and prompt configurations provide novel insights into leveraging LLM-derived knowledge for speech processing applications.
\end{abstract}
\section{Introduction}
\label{sec:intro}
% Why SSL for Speech fuoundational model?
Speech Foundational Models (SFMs) have received considerable research interest in recent years as they demonstrate state-of-the-art results in multiple datasets such as Whisper \cite{radford2023robust}, Canary \cite{puvvada24_interspeech}, USM \cite{zhang2023google}, and MMS \cite{mms}. These models are not only capable of processing multilingual speech, but also multiple tasks including Automatic Speech Recognition (ASR) and Automatic Speech Translation (AST).  
Unlike Natural Language Processing (NLP) tasks that require only text for training \cite{vas2017bert}, speech processing tasks require high-quality labeled speech data, which is less abundant compared to text-only data. 
In order to utilize data of various qualities, Semi-Supervised Learning (SSL) methods are then proposed in ASR \cite{zhang2020pushing, park20d_interspeech, hwang2022large, synnaeve2020endtoend} to train the model with pseudo labels. 
% which can address the shortage of supervised speech data by leveraging unlabeled audio for model training. 
% either select the unsupervised data \cite{lu22_interspeech} or enhance the quality of pseudo-labels \cite{chen2023improving, xi2024semi}. 
% However, other speech processing tasks such as AST are understudied. 

% LLM for correction
Several SSL studies have focused on optimization of the training strategy, such as developing multi-view or multi-objective loss to improve the training process \cite{cui2012multi, chung2020semi,niko2021gtc,khonglah2020issl}, and strengthening SSL models to enable more efficient label update \cite{higuchi2021momentum,higuchi2022momentum,zhu2023apl}. 
The selection strategy for unsupervised data is investigated in \cite{lu22_interspeech}, where contrastive selection methods are proposed to identify acoustically similar speech for a target domain. 
% The external text data is leveraged in \cite{li-vu-2024-improving} to align speech and text representations. 
Improving the quality of the pseudo-labels is another key focus in SSL. For example, in \cite{chen2023improving} a Language Model Filter is introduced to identify and refine high-quality pseudo-labels, particularly addressing domain mismatches within curated unsupervised datasets.
Large Language Model (LLM) is further explored in \cite{xi2024semi} to enhance the correction steps of monolingual speech in the code-switching ASR scenario. 
Different from ASR, which targets fixed transcriptions in the source language, AST deals with dynamic output labels that can vary across valid translations \cite{wang21r_interspeech}. 
While recent research has explored the use of LLMs for AST by developing advanced prompting strategies and modifications to model architectures \cite{chen-etal-2024-llast, hu2025chain, mundnich2025zero}, the SSL approaches remain understudied. 
% LLMs, trained on extensive multilingual text corpora, can naturally capture translation ambiguities, stylistic nuances, and cross-linguistic equivalencies. 
As translation hypothesis generated by SFMs is single and fixed, 
% Leveraging the probabilistic reasoning capabilities of LLMs offers a powerful approach to refine these pseudo-labels. 
LLMs can be served as a promising tool to produce context-aware, diverse, and valid AST pseudo-labels.

Moreover, existing SSL methods typically use relatively small student models and do not evaluate their performances on large-scale models or SFMs. 
% that tend to produce less accurate labels 
% , which can provide stronger baselines and more accurate labels.
Curated and well-prepared datasets are often used in an unsupervised manner by ignoring their ground truth labels, but they fail to reflect the complexity and diversity in real-world audios.
% In addition, most SSL techniques rely on curated or well-prepared datasets as their source of unsupervised data, disregarding ground truth labels. 
% A significant challenge remains in effectively harnessing in-the-wild data, which is more abundant and reflects a wider range of complex and diverse acoustic environments.
While SFMs can deliver more accurate and high-quality text hypothesis, leveraging in-the-wild audio which has inherent variability and noise still remains a challenge. 

In this work, we propose \emph{\textbf{L}}LM \emph{\textbf{E}}nhanced \emph{\textbf{S}}emi-\emph{\textbf{S}}upervised Learning (\emph{\textbf{LESS}}) method that aims to improve the quality of in-the-wild data for both ASR and AST. We make the following contributions:
\begin{itemize}
[itemsep=2pt,topsep=0pt,parsep=0pt]
    \item A novel approach that combines SSL techniques with LLMs for hypothesis correction is introduced, to improve pseudo-labels generated by state-of-the-art SFMs and address challenges posed by noisy and diverse audios directly sourced from YouTube \footnote{\url{youtube.com}}.  
    \item Experiments are carried out in Mandarin (ZH) ASR, achieving a significant 3.8\% Word Error Rate (WER) reduction on the WenetSpeech \cite{zhang2022wenetspeech} testset. The choice of LLMs, WER Prompting, and optimal filtering thresholds are discussed, offering new perspectives on applying LLMs to speech processing tasks. 
    \item We validate LESS framework on Spanish-to-English (ES-to-EN) AST, attaining state-of-the-art BLEU scores of 34.0 and 64.7 on the Callhome and Fisher \cite{fisher-callhome} test sets respectively, demonstrating LESS’s ability to generalize across languages, tasks, and domains.
    \item To ensure fully reproducibility, all SFMs and LLMs used are publicly available and the recipe is open-sourced \footnote{\url{github.com/nvidia-china-sae/mair-hub/tree/main/speech-llm/less\_recipe}}. 
\end{itemize}

\section{Methods}
\subsection{LESS pipeline}
\begin{figure}[h]
\centerline
{\includegraphics[width=4.9cm]{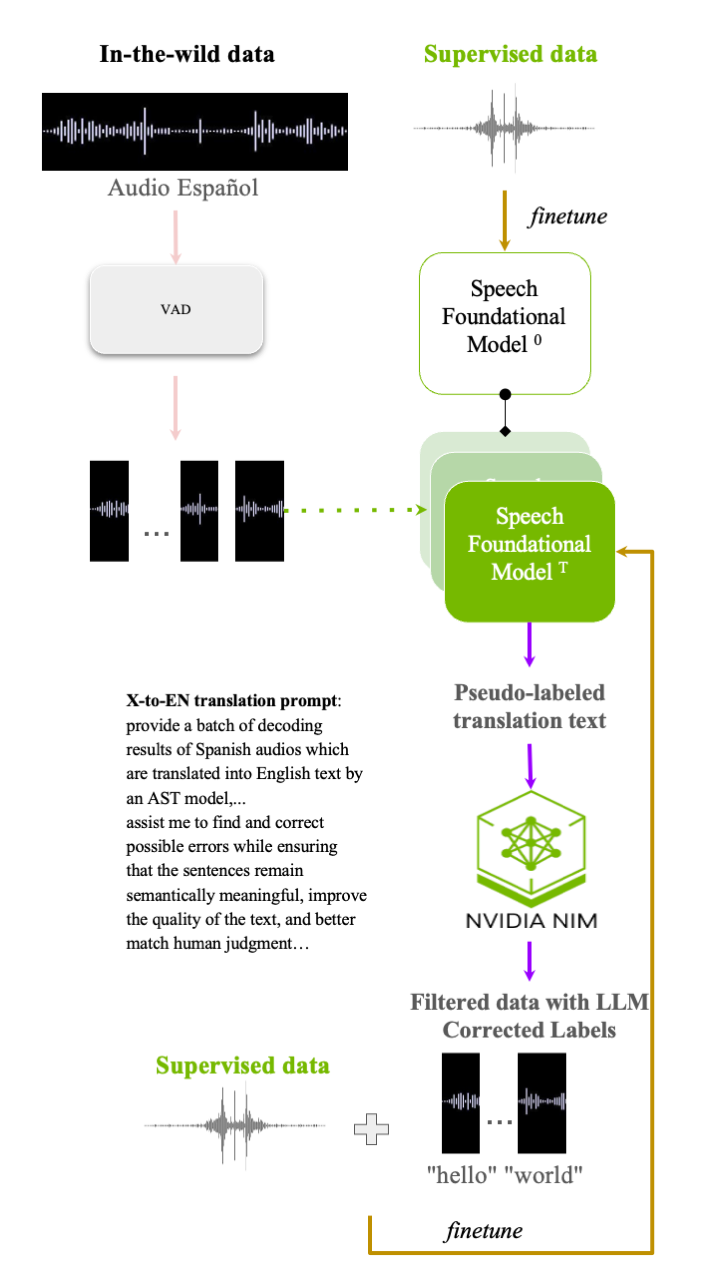}}
\caption{The proposed LESS pipeline is illustrated taking the ES-to-EN AST task as an example. The process begins by finetuning the SFM at T=0 with supervised Spanish data (orange arrow).
Unsupervised Spanish data collected from real-world sources are segmented by a VAD module. 
These Spanish segments are then transcribed into English using the initial SFM (SFM 0). The resulting English transcriptions are further refined by an LLM. 
After this, a data filtering step is applied. The filtered, LLM-corrected transcriptions are combined with the original supervised data to further finetune the SFM. This iterative process continues until convergence.}
\label{fig:overview_figure}
\end{figure}

The proposed pipeline is shown in Fig. \ref{fig:overview_figure}, which takes the ES-to-EN AST task and the Noisy Student Training (NST) method \cite{xie2020nxt} as an example. 
Initially, supervised Spanish data is employed to finetune the SFM, as indicated by the orange arrow in the figure. This finetuned SFM model serves as the initial seed (T=0) for the following iteration. 
Next, unlabeled Spanish raw audios collected from Youtube are segmented into shorter utterances using a Voice Activity Detection (VAD) module. The segmented audios are then processed by the seed model to produce English translations. 
Subsequently, a text-based LLM is requested to correct the English hypotheses, with the specific prompt details provided in Fig. \ref{fig:overview_figure}. 
After LLM correction, we apply the data filtering strategy described in \cite{chen2023improving, xi2024semi} that compares the WER of original hypothesis from greedy decoding and the LLM corrected result as a filtering threshold, to select high quality translations. 
The filtered, corrected pseudo-labeled data are then combined with the supervised data to further finetune the SFM, producing a new student model for the next iteration. 
This process is repeated iteratively until the model performance converges.

Furthermore, this pipeline can be easily adapted to other source languages. Only the input speeches need to be changed, while the remaining components require no modification. 
For the ASR task, the process closely follows that of AST, with changes limited to specifying the desired task of the SFM and updating the LLM prompt accordingly.

\subsection{In-the-wild data collection and processing} 
The unsupervised data are sourced directly from YouTube, under the Creative Commons license. The dataset comprises in-the-wild audio samples that are randomly chosen with a broad range of YouTube videos without any manual cleaning or pre-processing. This can capture real-world acoustic variability, noise, and other natural conditions without bias introduced by data curation. 
These raw audio files are then processed using the open-source Silero VAD \cite{SileroVAD} to detect and extract speech segments. The detected speech segments are then simply merged consecutively to produce final segments at maximum of 20 seconds in length, optimizing them for ASR and AST tasks.
As illustrated in Table \ref{table:data_all}, unsupervised 1590 hours of Mandarin data for ASR and 868 hours of Spanish data for AST are used in our studies. 

\section{Experiments} 
\label{section:data}
\subsection{Curated supervised data}
AISHELL1 \cite{aishell1} is an open-source Mandarin ASR dataset consisting of about 180 hours of speech. 
The Spanish Fisher and Callhome \cite{fisher-callhome} datasets contain 170 and 15 hours of telephony speech respectively, the English translation texts of which are used for ES-to-EN AST finetuning. The detailed duration is shown in Table \ref{table:data_all}. 

\subsection{Evaluation data}
In addition to evaluating performances on the test sets of AISHELL1, Fisher and Callhome, our study also reports results on the test sets of ZH AISHELL2 \cite{du2018aishell}, out-of-domain ZH Wenet Speech \cite{zhang2022wenetspeech} (referred to as wenet\_meeting), and Common Voice ES-to-EN AST \cite{ardila2020common} datasets.

\vspace{-10pt}
\begin{table}[htbp]
\centering
\caption{Duration of training data for curated ZH ASR and ES-to-EN AST, as well as in-the-wild YouTube data. }
\begin{resizebox}{0.8\columnwidth}{!} {\begin{tabular}{c  c  c}
\toprule
Language \& Task & Dataset & Duration (hours)\\
\midrule
\multirow{2}{*}{ZH ASR} & AISHELL1 & 180  \\
 & YouTube & 1590 \\
\midrule
\multirow{3}{*}{ES-to-EN AST} & Fisher & 170  \\
 & Callhome & 15 \\
& YouTube & 868 \\
\bottomrule
\end{tabular}
}\end{resizebox}
\label{table:data_all}
\end{table}

\vspace{-10pt}
\subsection{Experimental setup}
Whisper Large-v3 \footnote{\url{huggingface.co/openai/whisper-large-v3}}, one of the widely adopted open-source SFMs developed by OpenAI, is utilized in our experiments. It processes multilingual speech such as English, Mandarin and Spanish, and supports multiple tasks including ASR, AST and Language Identification. 
The Yi-Large \footnote{\url{build.nvidia.com/01-ai/yi-large}} LLM is employed in the ZH ASR task, and LLama-3 \footnote{\url{build.nvidia.com/meta/llama3-70b}} is used in the ES-to-EN AST experiments. 
All LLM requests are processed through NVIDIA NIM \footnote{\url{developer.nvidia.com/nim}}, which provides optimized microservices to deploy and serve various open-source models. 
% Epoch, GPU, lr, toolkit
The evaluation metrics are WER for ZH ASR and SacreBLEU score \cite{post-2018-call} for ES-to-EN AST accordingly, and the default value of filtering threshold is set to 0.1 for both ASR and AST \footnote{As BLEU Score is calculated among the whole datasets instead of per sentence, we use the WER as the threshold for AST as well.}. 
We finetune the Whisper-Large model for 5 epochs from scratch at each iteration with the learning rate 1e-5.
% and Deep Speed \cite{rasley2020deepspeed} is enabled during training. 
Model averaging is performed, and we use greedy decoding to simplify and speed up the inference pipeline. We adapt the weighted multiplexing for supervised and pseudo-labeled data, and all experiments are carried out using K2 Icefall toolkit
\cite{k2}.

\section{Results and Analysis}

\subsection{ZH ASR}
\subsubsection{Results}
The WER results for ZH ASR are presented in Table \ref{table:zh_asr_results} including the supervised baseline (Sup.) which is trained on the AISHELL1 dataset, and the standard NST which utilizes Mandarin YouTube data with pseudo-labels generated by the baseline model. 
Since in-the-wild data are randomly harvested from Youtube without any cleaning or selection as discussed in \ref{section:data}, WER performances remain comparable on AISHELL1 and AISHELL2 test sets, revealing that directly applying SSL to in-the-wild data is more challenging compared to other methods that use the curated or well-prepared datasets such as \cite{zhang2020pushing, chen2023improving}. 
However, the WER reduces in the wenet\_meeting, which is more related to the in-the-wild data and contains more noisy and diverse acoustics. 
Then, LESS is applied in subsequent experiments at each iteration of NST. 
Comparing the first NST round with and without LESS, an absolute 0.9\% of WER improvement is observed on the wenet\_meeting test set. 
After the three rounds of NST using LESS, WER consistently improves on the wenet\_meeting evaluation set, finally from 15.0\% to 13.9\%, amounting to an absolute reduction of 3.8\% compared to the Sup. baseline. Although there are limited  improvements on the AISHELL-1 and AISHELL-2 due to domain mismatch, the LESS method can benefit the model's robustness on a noisier and out-of-domain testset, demonstrating applying LESS in ASR tasks can have better generalization to out-of-domain evaluations. 

\vspace{-10pt}
\begin{table}[htbp]
\centering
\caption{ZH ASR WERs (\%) of using AISHELL1 as supervised dataset and YouTube as unsupervised dataset. Results include the supervised baseline, standard NST without LESS, and three iterations of NST with LESS. }
\begin{resizebox}{1.0\columnwidth}{!} {\begin{tabular}{ccccc}
\toprule
Model & LESS & AISHELL1 & AISHELL2 & wenet\_meeting\\
\midrule
Sup. & - & \textbf{2.9} & 5.3 & 17.7  \\
\midrule
Iter 1 & N & 3.0 & 5.3 & 15.9 \\
\midrule
Iter 1 & \multirow{3}{*}{Y} & 3.0 & 5.3 & 15.0 \\
Iter 2 &  & 3.0 & 5.3 & 14.2 \\
Iter 3 &  & 3.0 & \textbf{5.2} & \textbf{13.9} \\
\bottomrule
\end{tabular}
}\end{resizebox}
\label{table:zh_asr_results}
\end{table}
\vspace{-10pt}
\subsubsection{Discussions}
\textbf{LLM selection}. To measure how different LLMs impacts the proposed LESS method, we choose Qwen2.5-coder-32b-instruct \footnote{\url{build.nvidia.com/qwen/qwen2_5-coder-32b-instruct}} for a comparison with Yi-Large, with the same pseudo-labels from the first iteration of NST in ZH ASR training (Model A in Table \ref{table:llm_analysis}) and the same prompt. 
The Yi-Large model is trained with 3.1 trillion tokens of English and Chinese corpora and the training data of Qwen2.5-coder-32b-instruct includes 5.5 trillion tokens from source code, text-code grounding and synthetic data. 
Both LLMs report that they have been trained in Mandarin, making them well-suited in our experimental setup. 
We present results with the filtering threshold 1.0, which means that all the LLM-corrected hypotheses are directly used in NST without data filtering. This allows for a straightforward comparison with the baseline Model A. 
As shown in Table \ref{table:llm_analysis}, Yi-Large outperforms Qwen on the wenet\_meeting by achieving lower WERs (Model B vs. C). 
This may be due to the fact that the Qwen model has been post-trained primarily for coding tasks, which could limit its effectiveness in ASR hypothesis correction. 
In contrast, the Yi-Large model is a more general-purpose LLM, which appears to be better suited for this task. 
However, a slight performance drop in Model C compared to Model A is observed, possibly because the LLMs do not always refine hypotheses accurately and neither of the two open-source LLMs are post-trained on spoken language data.
% We believe that if the LLM is further trained with additional data, obtained from ASR or AST outputs and focused on text refinement or error correction, it could be beneficial for these tasks.
% Exploring this direction will be a key focus in our future work.

\textbf{WER Prompting}. Experiments are carried out with two LLM prompt variants: one directs the LLM to correct potential errors in the hypothesis, while the other instructs it to both perform the correction and calculate the WER between the original input hypothesis and the corrected version during generation process, which is referred to as \textit{WER Prompting}.
Model C and Model D in Table \ref{table:llm_analysis} show that the prompt with \textit{WER Prompting} can obtain a notable WER reduction on the wenet\_meeting, decreasing from 16.2\% to 15.8\%, and is better compared to Model A. 
Although the WER values generated by the LLM are often inaccurate and are not directly utilized in our experiments, including \textit{WER Prompting} proves to be more practical and can benefit downstream tasks\footnote{Due to page limitations, detailed prompt is provided in our recipe.}.  

\textbf{Data filtering}. We filter in-the-wild training data by calculating the WER between the initial greedy hypothesis and its LLM-refined version. We found that the stricter filtering threshold of 0.1 (Model E) yields a model with a better performance compared to the 1.0 threshold (Model D). As the LLM does not always generate a perfect correction, a proper filtering threshold is essential for our proposed LESS method. Therefore, we select the Yi-Large LLM with WER Prompting and a threshold of 0.1 for our primary ASR experiments.

\begin{table}[h]
\centering
\vspace{-10pt}
\caption{Results for various Mandarin LLMs including Qwen and Yi-large, evaluated under different configurations. \textit{WER Prompting} indicates whether the prompt instructs the LLM to generate the WER when correcting the hypotheses. All WERs (\%) are reported on the ZH wenet\_meeting dataset.}
\begin{resizebox}{1.0\columnwidth}{!} {\begin{tabular}{c c  c  c   c}
\toprule
Model & LLM	& \makecell{Filtering\\threshold} & \makecell{\textit{WER} \\ \textit{Prompting}} & wenet\_meeting \\
\midrule
A & - & - & - & 15.9 \\
\midrule
B & Qwen & 1.0 & N & 16.9  \\
% C &                      & 0.1 & N & 15.70 \\
\midrule
C & \multirow{3}{*}{Yi-Large}  & 1.0 & N  & 16.2 \\
D &                            & 1.0 & Y  & 15.8 \\
E &                            & 0.1 & Y & \textbf{15.0} \\
\bottomrule
\end{tabular}
}\end{resizebox}
\label{table:llm_analysis}
\end{table}

\vspace{-10pt}
\subsection{ES-to-EN AST}

Callhome and Fisher are combined to form the supervised dataset, with the YouTube data as unsupervised, the results of which are shown in Table \ref{table:es_ast_results}. 
% Under this experimental setup, a new baseline (Sup.) of using both supervised datasets to finetune the Whisper-Large model is established. 
For comparative purposes, we additionally report the ESPnet results \footnote{\label{espnet}\url{github.com/espnet/espnet/blob/master/egs2/fisher\_callhome\_spanish/}}, which indicates that our supervised baseline already exhibits a high level of performance.
We perform one iteration of standard NST with pseudo-labels generated from the Sup. baseline, where the BLEU score on Callhome and Fisher test sets slightly drop with directly adding pseudo-labeled in-the-wild data without LESS. This shows applying SSL to AST with in-the-wild data is challenging as well similar to ASR. 
Next, LESS is applied to the standard NST, and after one iteration of NST, using LESS clearly outperforms both the supervised baseline model and the standard NST model, achieving the highest BLEU scores in all evaluation datasets, with remarkable 34.0 and 64.7 in test sets of Callhome and Fisher, as well as the 37.3 in Common Voice test set. Due to limited time and resource, we conduct only one iteration of NST in this experiment. 

\vspace{-10pt}
\begin{table}[h]
\centering                                           \caption{Performances of ES-to-EN AST models trained with supervised Callhome-Fisher data, and in-the-wild Youtube as unsupervised data. Results from ESPnet are included for comparison. BLEU scores are reported in test sets of Callhome, Fisher, and Common Voice.}
\begin{resizebox}{1.0\columnwidth}{!} {\begin{tabular}{c  c  c c c c c c c}
\toprule
Model & LESS & Callhome & Fisher & Common Voice \\
\midrule
ESPnet\textsuperscript{[\ref{espnet}]} & - & 21.7 & 50.5 & -\\
\midrule
Sup. & - & 33.5 & 64.2 & 36.7\\
\midrule
% \cmidrule(lr){1-1}\cmidrule(lr){2-2}\cmidrule(lr){3-3}\cmidrule(lr){4-5}\cmidrule(lr){6-8}\cmidrule(lr){9-10}
\multirow{2}{*}{Iter 1} & N & 33.2 & 64.0 & 36.9\\
% \midrule
& Y & \textbf{34.0} & \textbf{64.7} & \textbf{37.3}\\
\bottomrule
\end{tabular}
}\end{resizebox}
\label{table:es_ast_results}
\end{table}
\vspace{-10pt}
We find that applying LESS to AST tasks yields consistent gains not only on in-domain test sets but also out-of-domain audios. We believe this improvement arises from the fact that LLMs are trained on large-scale multilingual text corpora, they can capture translation ambiguities and cross-linguistic equivalences. Leveraging this capability, LLMs can generate diverse pseudo-labels representing multiple valid translation possibilities, leading to more flexible and robust models across different languages and domains. 

\vspace{-10pt}
\section{Conclusions}
In this work, we present LESS for SFMs, which enhances the utilization of in-the-wild data by incorporating LLMs within a general NST framework.
% the hypothesis correction capabilities of LLMs within 
Experiments are conducted in the ZH ASR and ES-to-EN AST tasks, with 3.77\% WER reduction in ASR, and remarkable 34.0 and 64.7 BLEU scores in Callhome and Fisher AST test sets, revealing LESS's effectiveness across different languages, tasks as well as domains. 
Furthermore, various LLMs and prompt configurations facilitate a novel understanding of how to leverage LLM-derived knowledge for speech processing tasks. 
Looking ahead, we aim to strengthen the model's error correction capabilities through post-training on data that particularly source from ASR or AST outputs. 
Additionally, we plan to explore LESS to multi-modal LLMs that can process speech inputs and generate richer text contents. 
We also intend to apply this approach to a broader range of speech processing tasks, such as Spoken Question Answering.


\ninept
\bibliographystyle{IEEEbib}
\bibliography{strings,refs}

\end{document}